\documentclass[times,twocolumn,final,authoryear]{elsarticle}

\usepackage{framed,multirow}

\usepackage{amssymb}
\usepackage{latexsym}

\usepackage{graphicx}
\usepackage{tikz,pgfplots}
\usepackage{pgfplotstable}

\usetikzlibrary{tikzmark,shapes}
\usetikzlibrary{decorations.pathmorphing,decorations.pathreplacing,decorations.shapes}
\usetikzlibrary{matrix}
\usepgfplotslibrary{groupplots}
\usepackage{bm}
\usepackage{multirow}

\usepackage{amsthm}
\usepackage{xcolor}
\usepackage{soul}

\usepackage[small]{caption}
\usepackage{graphicx}
\usepackage{amssymb}
\usepackage{times,amsmath,float} 
\usepackage[lofdepth,lotdepth]{subfig}
\usepackage{enumitem}
\usepackage{epstopdf}
\usepackage{booktabs}

\newtheorem{remark}{Remark}[section]
\newtheorem{lemma}{Lemma}
\newtheorem{definition}{Definition}[section]

\newcommand{\lin}{\mathbf{x}}
\newcommand{\col}{\mathbf{y}}
\newcommand{\measa}{\boldsymbol{\alpha}}
\newcommand{\measb}{\boldsymbol{\beta}}
\newcommand{\measlin}{\boldsymbol{\kappa_r}}
\newcommand{\meascol}{\boldsymbol{\kappa_c}}
\newcommand{\coupl}{\boldsymbol{\gamma}}
\newcommand{\KL}{\mathrm{KL}}

\newcommand{\clustlin}{\mathbf{z_R}}
\newcommand{\clustcol}{\mathbf{z_C}}

\newcommand{\hatclustlin}{\mathbf{\hat{c}_L}}
\newcommand{\hatclustcol}{\mathbf{\hat{c}_C}}

\newcommand{\veca}{\mathbf{u}}
\newcommand{\vecb}{\mathbf{v}}

\newcommand{\vecc}{\mathbf{x}}

\newcommand{\matA}{\mathbf{A}}
\newcommand{\matM}{\mathbf{M}}

\newcommand{\CCOT}{\texttt{CCOT}}
\newcommand{\CCOTGW}{\texttt{CCOT-GW}}

\newcommand{\NMF}{\texttt{NMF}}

\newcommand{\ie}{i.e.}

\DeclareMathOperator*{\argmin}{\rm argmin}

\newcommand{\one}{\mathbf{1}}
\newcommand{\e}{\mathrm{e}}
\usepackage{url}

\definecolor{newcolor}{rgb}{.8,.349,.1}

\begin{document}

\thispagestyle{empty}
                                                             
\begin{frontmatter}

\title{Rank-one partitioning: formalization, illustrative examples, and a new cluster enhancing strategy}

\author{Charlotte Laclau, Franck Iutzeler, Ievgen Redko}

\address[1]{Univ Lyon, UJM-Saint-Etienne, CNRS,\\
        Institut d Optique Graduate School, Laboratoire Hubert Curien UMR 5516\\
        F-42023, Saint-Etienne, France}
\address[2]{Univ. Grenoble Alpes, CNRS, Grenoble INP, LJK\\
 38000 Grenoble, France}

\begin{abstract}
In this paper, we introduce and formalize a rank-one partitioning learning paradigm that unifies partitioning methods that proceed by summarizing a data set using a single vector that is further used to derive the final clustering partition. Using this unification as a starting point, we propose a novel algorithmic solution for the partitioning problem based on rank-one matrix factorization and denoising of piecewise constant signals. Finally, we propose an empirical demonstration of our findings and demonstrate the robustness of the proposed denoising step. We believe that our work provides a new point of view for several unsupervised learning techniques that helps to gain a deeper understanding about the general mechanisms of data partitioning.

\end{abstract}

\end{frontmatter}


\section{Introduction}
\label{sec1}
Cluster analysis aims to gather data instances into groups, called clusters, where instances within one group are similar among themselves while instances in different groups are as dissimilar as possible. Clustering methods have become more and more popular recently due to their ability to provide new insights into unlabeled data that may be difficult or even impossible to capture for a human being. Clustering methods are often categorized into two main frameworks, notably probabilistic and metric-based methods. Despite a large variety of existing approaches, several studies proved that sometimes seemingly distinct methods actually optimize the same objective function. This is the case, for instance, for k-means and EM for Gaussian mixture models clustering where the former can be shown to be a special case of the latter. On the other hand, different versions of non-negative matrix factorization \citep{lee99} minimize an objective function similar to a constrained k-means problem \citep{Ding:2010:CSM:1687044.1687110} and thus are intrinsically linked with probabilistic models. 

In this paper, we consider another unifying point of clustering algorithms that, in their vast majority, follow a two-step approach during which they (1) summarize the structural information about the data set at hand and (2) use this information to derive the final partitions. For the first step, such summarizing information can be given by an embedding into a lower dimensional space, e.g. matrix factorization methods, or by information given by one or more eigenvectors of the similarity matrix, as in spectral methods. In the latter case, one provably efficient approach is to extract the Fiedler vector of the Laplacian matrix calculated based on the similarity (or distance) matrix build from the data set at hand. In such case, one essentially uses a rank-one information that, after some post-processing, reveals all the necessary information about the partitioning of a high-dimensional data set. This observation is the core motivation of our work that prompts us to formalize such two-step learning strategy where the summarizing information from the first step is given by rank-one objects not necessarily related to spectrum of the data matrix. We consider several seemingly different methods and show that the rank-one summary provided by them to derive the final partition is extremely similar. Finally, we use the graph-theoretical interpretation of matrix factorization to show that this latter can also be seen as a valid rank-one partitioning method and couple it with a general signal denoising technique that enhances clustering information of the obtained decomposition. We would like to underline that our paper formalizes and presents the connection between the above-mentioned methods in an exploratory fashion: our foremost goal is to illustrate the intuition and provide a new point of view allowing to understand the link that exists between unsupervised learning approaches, that seem completely different at first sight. To this end, we further note our primary objective is not to obtain superior performance using our method but rather to provide a vivid demonstration of its connection to other algorithms that fall into the category of rank-one partitioning methods.
Our contributions are thus as follows:
\begin{itemize}
    \item[1)] \emph{A formal definition of a rank-one partitioning} with illustrative examples of well-known unsupervised learning algorithms that fall into the introduced category. We show that the results obtained using such methods are highly similar in practice despite the seeming differences between them. 
    \item[2)] \emph{A matrix factorization method for Step 1.} We provide a unifying view for different rank-one partitioning methods based on the loss function of the entropic regularized optimal transportation problem. We further use it to derive a simple rank-one matrix factorization approach for the introduced problem and provide a graph-theoretic justification of the soundness of the approach.
    \item[3)] \emph{An efficient method for Step 2.}  From cluster generating vectors, the construction of the actual clusters is often application-driven or done empirically in the literature. In order to be as general as possible, we consider the direct problem of denoising a signal taking a few number of (noisy) values. The proposed approach usually performs better as a plugin Step 2 for clustering methods of the literature.
\end{itemize}
The rest of this paper is organized as follows. In Section \ref{sec:rankone}, we present the rank one partitioning problem and its two-step formulation. In Section \ref{sec:proposed}, we first explore the relationship between the algorithms falling into the proposed framework and then propose a matrix factorization-based clustering method. In Section \ref{sec:clust}, we provide denoising scheme for the proposed method that enhances the block structure of the vectors resuming the data matrix. We evaluate our method and compare it to several other methods on both synthetic and real-world data in Section \ref{sec:EmpAn}. 

\section{Problem setup}
\label{sec:rankone}
In this section, we briefly introduce the notations used throughout the paper and then present our definition of rank-one partitioning learning.
\subsection{Notations}
In what follows, we denote the considered data matrix by $\matA \in \mathbb{R}^{m\times n}$ where $m$ denotes the number of instances and $n$ is the number of features. We use bold capital letters for matrices, e.g. $\matA$ and bold small letters for vectors, e.g. $\veca$. We denote the normalized Laplacian matrix of $\matA$ by $L(\matA):= I-\mathbf{D}^{-\frac{1}{2}}S_\matA\mathbf{D}^{-\frac{1}{2}}$ where $S_\matA \in \mathbb{R}^{m \times m}$ is a similarity matrix calculated from $\matA$ (when $m=n$, we let $S_\matA = \matA$, i.e., we suppose that $\matA$ is an adjacency matrix of some graph) and $\mathbf{D} = \text{diag}(\sum_{j=1}^n (S_\matA)_{ij})$. We write $E_\varepsilon(\matA)$ to denote the eigenspace of $\matA$ and suppose that eigenvalues are sorted in the increasing order with respect to their value, i.e., for any $i<j, \lambda_i > \lambda_j$.



\subsection{Rank-one partitioning}
\label{sec:def}
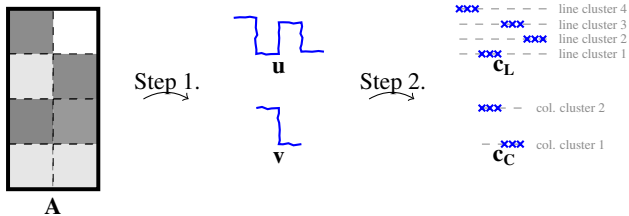
\begin{figure}[!h]
    \centering
\begin{tikzpicture}[scale=0.6]

\coordinate (M1) at (0,4) ;
\coordinate (M2) at (1,4) ;
\coordinate (M3) at (2,4) ;

\path (M1) + (0,-1) coordinate (M4);
\path (M2) + (0,-1) coordinate (M5);
\path (M3) + (0,-1) coordinate (M6);

\path (M1) + (0,-2) coordinate (M7);
\path (M2) + (0,-2) coordinate (M8);
\path (M3) + (0,-2) coordinate (M9);

\path (M1) + (0,-3) coordinate (M10);
\path (M2) + (0,-3) coordinate (M11);
\path (M3) + (0,-3) coordinate (M12);

\path (M1) + (0,-4) coordinate (M13);
\path (M2) + (0,-4) coordinate (M14);
\path (M3) + (0,-4) coordinate (M15);

\draw [black,dashed,fill=gray!90] (M1) -- (M2) -- (M5) -- (M4) -- cycle ;
\draw [black,dashed,fill=gray!20] (M4) -- (M5) -- (M8) -- (M7) -- cycle ;
\draw [black,dashed,fill=gray!90] (M5) -- (M6) -- (M9) -- (M8) -- cycle ;

\draw [black,dashed,fill=gray!95] (M7) -- (M8) -- (M11) -- (M10) -- cycle ;
\draw [black,dashed,fill=gray!80] (M8) -- (M9) -- (M12) -- (M11) -- cycle ;
\draw [black,dashed,fill=gray!20] (M10) -- (M11) -- (M14) -- (M13) -- cycle ;
\draw [black,dashed,fill=gray!20] (M11) -- (M12) -- (M15) -- (M14) -- cycle ;

\draw [black,line width = 1.5pt] (M1) -- (M3) -- (M15) -- (M13) -- cycle   ;

\draw (M14) node[below] {\small $\mathbf{A}$};

\draw [->] (3,2) to [out=30,in=150] (4,2);
\draw (3.53,2.35) node {\small Step 1.};


\draw[blue, thick,decorate, decoration={random steps,segment length=2pt,amplitude=0.5pt} ] (5,3.8) -- (5.5,3.8) -- (5.5,3.0) -- (6,3.0) -- (6,3.7) -- (6.5,3.7) -- (6.5,3.05) -- (7,3.05); 
\draw (6,2.7) node {\small $\mathbf{u}$};

\draw[blue, thick,decorate, decoration={random steps,segment length=2pt,amplitude=0.5pt} ]   (5.5,1.8) -- (6,1.8) -- (6,1) -- (6.5,1) ; 
\draw (6,0.7) node {\small $\mathbf{v}$};

\draw [->] (8,2) to [out=30,in=150] (9,2);
\draw (8.53,2.35) node {\small  Step 2.};

\draw[dashed,color=gray!90] (10,4) -- (12,4) node[right] {\tiny line cluster 4};
\draw[dashed,color=gray!90] (10,3.66) -- (12,3.66) node[right] {\tiny line cluster 3};
\draw[dashed,color=gray!90] (10,3.33) -- (12,3.33) node[right] {\tiny line cluster 2};
\draw[dashed,color=gray!90] (10,3.0) -- (12,3.0) node[right] {\tiny line cluster 1};
\draw[blue, thick,decorate, decoration={crosses,segment length=3pt} ] (10,4) -- (10.5,4) ; 
\draw[blue, thick,decorate, decoration={crosses,segment length=3pt} ] (10.5,3.0) -- (11,3.0) ; 
\draw[blue, thick,decorate, decoration={crosses,segment length=3pt} ]  (11,3.66) -- (11.5,3.66); 
\draw[blue, thick,decorate, decoration={crosses,segment length=3pt} ] (11.5,3.33) -- (12,3.33); 
\draw (11,2.7) node {\small $\mathbf{c_L}$};

\draw[dashed,color=gray!90] (10.5,1.8) -- (11.5,1.8) node[right] {\tiny col. cluster 2};
\draw[dashed,color=gray!90] (10.5,1.0) -- (11.5,1.0) node[right] {\tiny col. cluster 1};
\draw[blue, thick,decorate, decoration={crosses, segment length=3pt} ]   (10.5,1.8) -- (11,1.8) ; 
\draw[blue, thick,decorate, decoration={crosses, segment length=3pt} ]    (11,1) -- (11.5,1) ; 
\draw (11,0.7) node {\small  $\mathbf{c_C}$};

\end{tikzpicture}
    \caption{Illustration of rank-one co-clustering based on a two-stage procedure. }
    \label{fig:coclust}
\end{figure}

We now formally defining the rank-one partitioning learner as a two-step learning procedure. 
\begin{definition}
Given an $m\times n$ matrix $\matA$, rank-one partitioning learner $\mathcal{A}:\mathbb{R}^{m\times n} \rightarrow \tilde{\mathbb{Z}}^m$ is defined as follows:
\begin{align*}
    \mathcal{A} = \{g\circ f|f:\mathbb{R}^{m\times n} \rightarrow \mathbb{R}^{m}, g:\mathbb{R}^{m} \rightarrow \tilde{\mathbb{Z}}^m, |\tilde{\mathbb{Z}}^m_\neq| = k, k\leq m\},
\end{align*}
where $\tilde{\mathbb{Z}}^m_\neq$ denotes the subset containing distinct values of $\tilde{\mathbb{Z}}^m$ given by $m$-dimensional vectors with integer elements.
\end{definition}
\begin{remark}
This definition trivially extends to the case of co-clustering when one seeks to find cluster-generating vectors of both rows and columns of a data matrix. In this case, the rank-one co-clustering learner is defined as follows:
\begin{align*}
    \mathcal{A} = \{g\circ f|f: &\mathbb{R}^{m\times n} \rightarrow \mathbb{R}^{m}\times\mathbb{R}^{n}, g:\mathbb{R}^{m}\times\mathbb{R}^{n} \rightarrow \tilde{\mathbb{Z}}^m\times \tilde{\mathbb{Z}}^n,\\
        &|\tilde{\mathbb{Z}}^m_\neq| = k_r, |\tilde{\mathbb{Z}}^n_\neq| = k_c, k_r\leq m, k_c\leq n\}.
\end{align*}
\end{remark}
In what follows, we consider a more general case of rank-one co-clustering where the cluster-generating vectors are extracted separately for rows and columns in case of clustering methods or jointly in case of co-clustering ones. The two-step approach underlying the composition of functions that $\mathcal{A}$ seeks for is illustrated by Figure~\ref{fig:coclust} and consists in: 
\begin{itemize}[itemindent=2em]
    \item[\textbf{Step 1}.] Learning a map $f$ that for any data matrix returns cluster-generating vectors $\veca\in\mathbb{R}^m, \vecb\in\mathbb{R}^n$ subsuming the block-structure of $\matA$. These vectors have no-block structure but the histograms of their values should contain a reduced number of (noisy) modes.
\end{itemize} 
This step can be seen as a particular case of dimensionality reduction algorithms with the rank fixed to 1 (see e.g. \citep[Section 5]{Collins:2001}). However, in the dimension reduction literature, rank-one approximations are usually considered as illustrative examples and not as valid learning strategies. 
\begin{itemize}[itemindent=2em]
    \item[\textbf{Step 2}.] Coordinate clustering of vector $\veca$ with a map $g$ that returns $\clustlin \in \tilde{\mathbb{Z}}^m$ with a finite number of values corresponding to the rows partition as explained above. The same process is repeated for $\vecb$ to obtain a column partitioning $\clustcol \in \tilde{\mathbb{Z}}^n$.
\end{itemize} 
We note that this part is often performed in an ad-hoc manner in the literature, depending on the application and the tools used in the first step. 

We now proceed to a presentation of different methods that proceed following such a two-stage procedure. 


\subsection{Examples of rank-one partitioning methods}
\label{subsec:examples}
We distinguish two principal approaches that can be seen as rank-one partitioning learners: spectral and statistical methods. Below, we present algorithms that belong to each of these categories.

\textit{Spectral methods}$\hphantom{1}$ As mentioned in the introduction, many spectral-based methods fall into the introduced framework of rank-one partitioning. Such methods can be roughly summarized as follows: (\textbf{Step 1}) pre-processing the data matrix and calculating a particular eigenvector associated with it; (\textbf{Step 2}) detecting distinct values in the extracted eigenvector. As for the first step, the pre-processing usually consists in calculating a similarity matrix (or adjacency matrix in case when the data matrix is related to a graph) and, optionally, scaling its rows and/or columns to some predescribed value. Then, three options have been explored in the literature: 
\begin{enumerate}
    \item Extracting the eigenvector of the Laplacian matrix associated with the second smallest eigenvalue \citep{Shi:2000:NCI:351581.351611,DingHZ01,Luxburg:2007:TSC:1288822.1288832}, \ie, 
        $$f(\matA) = \veca, s.t.\ L(\matA)\veca = \lambda_2\veca.$$ 
    The efficiency of such approach is justified by showing that this eigenvector has a sign-pattern allowing to bi-partition vertices of a graph in an optimal way \citep{fiedler73};
    \item Extracting the unit dominant right eigenvector of the scaled adjacency \citep{Pageetal98,Gorrec2019} matrix , \ie, 
        $$f(\matA) = \veca, s.t.\ \widetilde{\matA}\veca = \lambda_{n}\veca, \sum_{i=1}^{m}\widetilde{\matA}_{ij} = 1,\ \forall j.$$ 
    This method is essentially the reverse of the previous one as Fiedler vector of normalized Laplacian matrix and first non-trivial dominant eigenvector of random walk adjacency matrix are known to be related \citep{Meila01learningsegmentation}. A noticeable example of such approach is the famous \textsc{PageRank} algorithm. 
\end{enumerate}
As for the second step, all such methods then use the obtained eigenvector to perform jump detection, \ie, 
$$g(\veca) = \clustlin, s.t.\ \forall i,j, \clustlin_i = \clustlin_j \text{ if }|\veca_i - \veca_j|\leq \epsilon$$
with $\epsilon$ being a threshold defined by the user or ranking in order to partition the data and determine the final clusters. A column partitioning can be obtained similarly by looking at the left eigenvectors.

\textit{Statistical methods}$\hphantom{1}$ A different approach that can be seen as rank-one partitioning is based on statistical inference where the marginal distributions of rows are considered as cluster-generating vectors. One such method, introduced in \citep{Snijders1997} and further extended in \citep{channarond2012} and \citep{brault_17} under the name \emph{Largest Gaps}, aims at finding homogeneous groups of nodes in the adjacency matrix $\matA \in \mathbb{R}^{m \times m}$ through a two-step procedure. At \textbf{Step 1}, their approach computes the degrees of nodes and sorts them in the ascending order producing a vector $f(\matA) = \veca, s.t.\ \forall i = 1, \dots, m,\ : \ \veca_i = \sum_{j}\matA_{ij}/m$. The intuition behind this process was to consider that the elements of a given data matrix may be somehow proportional to the joint distribution of rows and columns and thus the row sums can be expected to carry some information about their corresponding marginal distribution.

This last observation is also at the core of the \CCOT~method proposed in \citep{laclau_17} where the authors seek to estimate a joint distribution $\coupl$ between the empirical measures defined as $ \measlin := \sum_{i=1}^m \delta_{\lin_i}/m ~~ \text{ and } ~~ \meascol := \sum_{i=1}^n \delta_{\col_i}/n$ where  $\{\lin_i \in \mathbb{R}^{n} \}_{i=1}^m$ and $\{\col_i \in \mathbb{R}^{m} \}_{i=1}^n$ are the rows and columns of the matrix $\matA$, respectively. This was done by solving the entropy regularized optimal transport problem \citep{conf/nips/Cuturi13} leading to the solution:
\begin{align}
   \coupl_{\varepsilon}^\star (\measlin,\meascol) =  \text{diag}(\veca) \e^{-\matM/\varepsilon}\text{diag}(\vecb).
   \label{eq:ccot}
\end{align}
where $\matM$ is a cost matrix, \ie, $M_{i,j} = d(\lin_i,\col_j)$ for some distance $d$\footnote{ $M_{i,j} = d(\lin_i,\col_j)  = \|\lin_i - \col_j\|$ if $\mathbf{A}$ is a square matrix. Otherwise, Gromov-Wasserstein transportation is used.}. As $\coupl$ is a valid joint distribution, it can be further factorized as $q(\hat{x},\hat{y}) = q(x) q(\hat{x}, \hat{y} \vert x,y) q(y),$ for two latent variables $\hat{x},\hat{y}$ so that vector $\veca$ can be seen as a marginal distribution of rows. The mapping $f$ can be then defined as:
\begin{align*}
    f(\matA) &= \veca, s.t.\ \coupl_{\varepsilon}^\star =  \text{diag}(\veca) \e^{-\matM/\varepsilon}\text{diag}(\vecb),\\
    &\forall i,j \sum_{i=1}^{m}(\coupl_{\varepsilon}^\star)_{ij} = 1/n, \sum_{j=1}^{n}(\coupl_{\varepsilon}^\star)_{ij} = 1/m.  
\end{align*}
At \textbf{Step 2}, the \emph{Largest Gaps} algorithm and its variations define the jumps in the obtained vector using a threshold that allows to split it into a certain number of homogeneous groups with a function $g(\veca)$ identical to that of spectral methods. \CCOT\ method, however, uses a more elaborate multi-scale denoising technique that allows to detect jumps in the obtained vector \citep{MatMei}. Contrary to all methods considered before, \CCOT\ also learns cluster-generating vectors for both rows and columns simultaneously and thus is directly suitable for clustering both modes of data. 

In the following section, we draw the parallels between all methods described above and show that simple rank-one matrix factorization can be seen as a combination of both spectral and statistical methods.

\section{Learning cluster-generating vectors with matrix factorization}
\label{sec:proposed}
The methods presented above illustrate two general trends for matrix partitioning: 1) the ones based on spectral information, such as \textsc{PageRank} and 2) the ones based on statistical inference, such as \emph{Largest Gaps}. Even though \CCOT\ was considered as a statistical method, it actually combines these two approaches as it aims at finding a joint probability distribution between lines and columns while enforcing a certain spectral structure due to the entropic regularization of the transport problem. Using this remark as a starting point, we further show that this connection between optimal transport and matrix factorization that can be used to retrieve rank-one information about the data instances and features of a given data matrix.

\subsection{A unifying loss: entropy-regularized transportation}
Let us consider the objective function related to an optimal transportation problem with entropic regularization\footnote{Due to space restrictions, we defer details about optimal transport to \ref{sec:ot}.}:
\begin{align}
\label{eq:fun}
   \mathcal{L}(\coupl ; \matM  ;\mathbf{a},\mathbf{b}) :=  \langle \matM , \coupl \rangle_F + \varepsilon ~ \KL(\coupl || \mathbf{a}\mathbf{b}^\top)
\end{align}
where $\coupl \in \mathrm{R}^{m\times n}$ is the coupling between lines and columns; $\matM\in \mathrm{R}^{m\times n}$ is the cost of moving from lines to columns, $\mathbf{a}\in \mathrm{R}^{m}$ and $\mathbf{b}\in \mathrm{R}^{n}$ are the weights of the lines and the columns, respectively.

The optimization problem defined in \eqref{eq:fun} can be shown to be linked to both spectral and statistical methods described above as follows: 
\begin{itemize}
    \item \emph{Spectral methods} As mentioned earlier, the solution of \eqref{eq:fun} is given by two scaling vectors obtained by applying the Sinkhorn-Knopp algorithm to the matrix $\e^{-\matM/\varepsilon}$. To this end, we note that the link between Sinkhorn-Knopp scaling vectors and \textsc{PageRank}'s leading eigenvector has already been mentioned by \citep{knight_08} even though it was not explicitly proved. The rationale behind this is to notice that \textsc{PageRank} algorithm essentially looks for the stationary distribution associated with the column stochastic adjacency matrix while Sinkhorn-Knopp algorithm seeks for scaling vectors that allow to obtain a doubly stochastic adjacency matrix with uniform stationary distribution. In this case, it is reasonable to assume that the scaling vectors reflect the highest contribution of each node to ``uniformizing" the stationary distribution just as the elements of the leading eigenvector found by \textsc{PageRank} identify the most highly influential nodes in the unbalanced stationary distribution. Finally, as the leading eigenvector of the (scaled) adjacency matrix carries similar information compared to the Fiedler vector used by other spectral methods, the link between \eqref{eq:fun} and these latter follows as well. 
    
    \item \emph{Statistical methods} If one takes a uniform transport cost $\matM = \one \one^\top$ and defines $ \mathbf{a}_i = \sum_{j=1}^n \matA_{i,j}$ and $\mathbf{b}_j =  \sum_{i=1}^m \matA_{i,j}$, then the dual vectors obtained by Sinkhorn's algorithm corresponds to an extension of the \textit{Largest Gaps} method for rectangular matrices. As for \CCOT, $\mathcal{L}$ is minimized with respect to the coupling $\coupl$ with $\mathbf{a}=\one_m/m$ and $\mathbf{b}=\one_n/n$ and  $\matM$ defined as described in \ref{subsec:examples}. However, instead of using the optimal coupling $\coupl$, the authors use the dual scaling vectors $\veca$ and $\vecb$ produced by Sinkhorn's algorithm.
\end{itemize}

\subsection{Proposed approach: rank-one matrix factorization} 
For a given data matrix $\matA$, we propose to consider the following minimization problem:
\begin{align*}
   (\veca,\vecb) &:=  \arg\min_{ (\veca,\vecb) } \mathcal{L}(\matA; \matM ; \veca,\vecb ) 
\end{align*}
where contrary to the original problem the minimization is performed over $\veca$ and $\vecb$. Note that contrary to \eqref{eq:fun} where $\mathbf{a}$ and $\mathbf{b}$ were fixed, here we aim to use their learned approximations as $\veca$ and $\vecb$ thus motivating such a change of notation. The introduced problem simply amounts to the minimization of $\KL(\coupl || \veca\vecb^\top)$ with $\coupl = \matA$ that does not necessarily represent some joint distribution. Instead, this problem aims to find two vectors $\veca$ and $\vecb$ that have sufficient entropy or small enough mutual information w.r.t. the matrix of interactions $\matA$. 

In the general case of a matrix $\matA\in\mathbb{R}^{m\times n}$ with potentially missing values, the calculation of $ \veca$ and $\vecb$ amounts to solving the following matrix factorization problem:
\begin{align*}
    \min_{(\veca,\vecb)\in\mathbb{R}^m\times \mathbb{R}^n} \sum_{\underset{ \matA_{i,j} \text{ exists } }{i,j=1}}^{m,n} \matA_{i,j} \log\left( \frac{\matA_{i,j}}{u_i v_j} \right) - \matA_{i,j} + u_i v_j .
\end{align*}
In what follows, we use non-negative matrix factorization to optimize this objective function following the algorithm proposed by \citep{lee99}. Note that the Kullback-Leibler divergence can be seen as a special case of the $\beta$-divergence (with  $\beta=1$). This means that the same reasoning may be used for  other $\beta$-divergences for both the regularization term of the optimal transportation problem and the matrix factorization loss function derived from it (see \citep{fevotte2011algorithms} for an account of different $\beta$-divergences applications in matrix factorization). Notably, for $\beta=2$, we obtain the Euclidean loss that has been used in regularized optimal transport and is widely used in matrix factorization, notably in recommender systems \citep{Koren:2009:MFT:1608565.1608614}).

It is worth noticing that the rank-one matrix factorization introduced above have not only statistical interpretation via its link to the entropy regularized optimal transport but also a spectral one. This follows from \citep[Proposition 4.1]{mirzal11} showing that optimizing the NMF objective function amounts to applying the relaxed ratio association to the item and the feature graph simultaneously where the graph is constructed using the data matrix $\matA$. The partitions in rank-$k$ factorization with $k>1$ are obtained by determining the largest projection on the axis of the rank-$k$ subspace. In case of rank-one decomposition, however, our intuition would be to recover the final partition by considering the magnitudes of projection on the only learned subspace. 

\begin{remark}
The connection between optimal transport and matrix factorization can also be shown using a different point of view. When $\mathbf{a}$ and $\mathbf{b}$ are defined as uniform vectors (\ie, as in the definition of $\measlin,\meascol$ in \CCOT), the Kullback-Leibler divergence term in (\ref{eq:fun}) becomes:
\begin{align*}
	\KL(\coupl || \mathbf{a}\mathbf{b}^\top) = \KL(\coupl || \bm{1} \bm{1}^T/mn). 
\end{align*}
Based on this expression, one may further note that as $ \varepsilon$ grows, the solution of the regularized optimal transport becomes closer to the uniform distribution and thus its factorization becomes:
\begin{align*}
	\bm{1} \bm{1}^T/mn \underset{\varepsilon \rightarrow \infty}{\sim} \coupl_{\varepsilon}^\star (\measlin,\meascol)  & =  \text{diag}(\veca ) \e^{-\matM/\varepsilon}\text{diag}(\vecb) 	\\
				(1/\veca) (1/\vecb)^T &\simeq  \e^{-\matM/\varepsilon} mn .
\end{align*}
Note that the obtained equation can be equivalently seen as a rank-one matrix factorization of the Gibbs kernel  $\e^{-\matM/\varepsilon}$ that depicts the joint interactions between two sets of objects underlying $\measlin$ and $\meascol$ (\ie, the lines and columns in \CCOT). On the other hand, in the extreme case when $\varepsilon \rightarrow \infty$, the Gibbs kernel $\e^{-\matM/\varepsilon}$ becomes a matrix of ones leading to a degenerate solution where $\veca$ and $\vecb$ are vectors of ones. 
\end{remark}

\subsection{Illustration} 
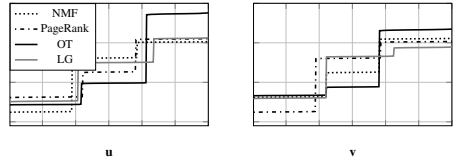
\begin{figure}[!t]
\centering
 \begin{tabular}{cc}
\begin{tikzpicture}[scale=0.5]

\begin{axis}[ 
 width=0.85\columnwidth, 
 height=0.6\columnwidth, 
 xmajorgrids, 
 yminorticks=true, 
 ymajorgrids, 
 yminorgrids,
 xlabel = {$\veca$},
 ymin = 0.0019,
 ymax=0.005,
 xmin = 1,
 xmax = 300,
   xticklabels=\empty,
  yticklabels=\empty,
  scaled y ticks = false,
legend style={at={(0,1)},anchor=north west, font=\small},
tick label style={font=\large}
 ]

 \addplot  [color=black,
                dotted,
                mark=none,
                mark options={solid},
                line width=1.2pt]  file {u_nmf_1.txt }; 
 \addlegendentry{NMF};

  \addplot  [color=black,
                dash pattern=on 1pt off 3pt on 3pt off 3pt,
                mark=none,
                mark options={solid},
                line width=1.2pt]  file {u_pg.txt}; 
 \addlegendentry{PageRank};

  \addplot  [color=black,
                mark=none,
                mark options={solid},
                line width=1.2pt]  file {u_sink.txt}; 
 \addlegendentry{OT};
 
  \addplot  [color=gray,
                mark=none,
                mark options={solid},
                line width=1.2pt]  file {v_marg.txt}; 
 \addlegendentry{LG};

\end{axis}
\end{tikzpicture}&
\begin{tikzpicture}[scale=0.5]

\begin{axis}[ 
 width=0.85\columnwidth, 
 height=0.6\columnwidth, 
 xmajorgrids, 
 yminorticks=true, 
 ymajorgrids, 
 yminorgrids,
 xlabel = {$\vecb$},
 ymin = 0.0019,
 ymax=0.005,
 xmin = 1,
 xmax = 300,
   xticklabels=\empty,
  yticklabels=\empty,
  scaled y ticks = false,
legend style={at={(0.36,1)},anchor=north east, font=\small},
tick label style={font=\large}
 ]

 \addplot  [color=black,
                dotted,
                mark=none,
                mark options={solid},
                line width=1.2pt]  file {v_nmf_1.txt }; 

  \addplot  [color=black,
                dash pattern=on 1pt off 3pt on 3pt off 3pt,
                mark=none,
                mark options={solid},
                line width=1.2pt]  file {v_pg.txt}; 

  \addplot  [color=black,
                mark=none,
                mark options={solid},
                line width=1.2pt]  file {v_sink.txt}; 
 
   \addplot  [color=gray,
                mark=none,
                mark options={solid},
                line width=1.2pt]  file {u_marg.txt}; 

\end{axis}
\end{tikzpicture}
\\
\end{tabular}
\caption{Comparison of the vectors obtained with PageRank, Matrix Factorization (NMF), CCOT (OT) and Largest Gaps (LG) on a simulated squared data matrix with a $3\times 3$ block structure.}
\label{fig:all_vectors}
\end{figure}

In the beginning of this section, we argued that some popular algorithms can be seen as instances of the introduced rank-one partitioning paradigm. We further showed that some of them can be equivalently expressed as a rank-one matrix factorization problem. In order to illustrate the links between these seemingly different methods, we show in 
Figure \ref{fig:all_vectors} the cluster-generated vectors obtained by them on the same synthetic data set. The simulated matrix contains 60 variables and 60 instances, and presents a block structure with 3 homogeneous groups, or clusters, for both instances and variables. The compared methods include the \CCOT, \textsc{PageRank}, the rank-one non-negative matrix factorization with Kullback-Leibler loss (\NMF) proposed above and the \textit{Largest Gaps} (LG) method. From this figure, we can see that all cluster-generated vectors present similar changes in slope, at close but slightly different locations, revealing the same type of steps for both $\veca$ and $\vecb$.

\section{A Novel Cluster-enhancing Procedure}
\label{sec:clust}

In the previous section, we presented several methods that can be used to compute the cluster-generating vectors $\veca$ and $\vecb$ of the Step 1 in the rank-one partitioning framework. 
Based on the nature of the obtained vectors, two possible cases are to be considered for Step 2: in the first one, when the blocks in $\veca$ and $\vecb$ are contiguous, these vectors can be seen as piece-wise constant noisy signals that should be denoised to obtain the clustering partitions; in the second case, when the blocks are non-contiguous, the clustering of the vector coordinates to a reduced number of values necessitates a preliminary sorting. The difference between the two scenarios is illustrated in Fig.~\ref{fig:potts}(a,b).

In this section, we introduce a general procedure that can be used to map coordinates of the obtained cluster-generating vectors to final clustering partitions. To this end, we start with a presentation of common methods for the denoising of 1D piece-wise constant signals. Then, we provide an original extension of these methods that allows us to efficiently define clusters from cluster-generating vectors from Step 2 of the rank-one partitioning framework. 

\subsection{Denoising of Piece-wise Constant Signals}

The problem of denoising piece-wise constant signals -- which corresponds to 1D-clustering -- has received a lot of attention in the literature \citep{basseville1993detection,little2011generalized}. As in unsupervised setting the number of clusters is unknown, we further concentrate our attention on two popular methods based on the Potts problem (see the recent papers of \citep{weinmann20151} and references therein). In general, for a given size-$n$ vector $\veca$, both of these methods aim at finding a piece-wise constant vector $\vecc$ by solving an optimization problem composed of two terms:
\begin{enumerate}
    \item[-] A fitting term taken as the $\ell_p$ norm $\|\vecc - \veca\|_p^p$ for some $p\geq 1$.
    \item[-] A regularization term, controlled by an hyper-parameter $\lambda_{\text{reg}}>0$, that penalizes the increments:
$$\left[x_2 - x_1,x_3 - x_2,\dots,x_{n} - x_{n-1} \right]^T := D\vecc$$
where $D \in \mathbb{R}^{n\times n-1}$ is the matrix such that $[D\vecc]_{i} = (x_{i+1}-x_{i})$ for $ i = 1,\dots,n-1$.
\end{enumerate}
 
Based on this general form, the $\ell_p$-Potts method is obtained by adding a unit penalty when one of the coordinates of the vector of increments is non-null (\ie, two consecutive coordinates of $\vecc$ are not equal). Using the $\ell_0$ semi-norm (i.e. the number of non-null entries) as a penalty for non-null increments, this problem can be used to obtain $\clustlin$ as follows:
\begin{align}
\tag{$\ell_p$--Potts}
\clustlin = \arg\min_{\vecc \in \mathbb{R}^n}  \|\vecc-\veca\|^p_p  +  \lambda_{\text{reg}} \|D\vecc\|_0 .
\end{align}

\begin{figure}[!t]
\centering

\begin{tikzpicture}[scale=0.4]
\begin{groupplot}[group style={group name=plotP, group size= 2 by 2},width=\textwidth]

\nextgroupplot[ 
 width=\columnwidth, 
 height=0.6\columnwidth, 
 xmajorgrids, 
 yminorticks=true, 
 ymajorgrids, 
 yminorgrids,
  xticklabels=\empty,
  yticklabels=\empty,
  xmin=0,
  xmax = 100,
  title={(a) Contiguous clustering problem and solution with  $\ell_1$-Potts},
tick label style={font=\large}
 ]

 \addplot  [color=black,
                only marks,
                mark=square*,
                mark size = 0.5pt,
                mark options={solid},
                line width=1.2pt]  file {Potts/contiguous_orig.txt }; 
\label{plotP:GT}

 \addplot  [color=blue,
                only marks,
                mark=star,
                mark size = 1.5pt,
                mark options={solid},
                line width=1.2pt]  file {Potts/contiguous_noise.txt }; 
\label{plotP:vec}

  \addplot  [color=red,
                only marks,
                mark=-,
                mark size = 1.8pt,
                opacity=0.7,
                line width=2.5pt]  file {Potts/contiguous_potts.txt }; 
\label{plotP:P}

\nextgroupplot[ 
 width=\columnwidth, 
 height=0.6\columnwidth, 
 xmajorgrids, 
 yminorticks=true, 
 ymajorgrids, 
 yminorgrids,
  xticklabels=\empty,
  yticklabels=\empty,
  xmin=0,
  xmax = 100,
 title={(b) Non-contiguous clustering problem},
tick label style={font=\large}
 ]

 \addplot  [color=black,
                only marks,
                mark=square*,
                mark size = 0.5pt,
                mark options={solid},
                line width=1.2pt]  file {Potts/unsorted_orig.txt };

 \addplot  [color=blue,
                only marks,
                mark=star,
                mark size = 1.5pt,
                mark options={solid},
                line width=2.5pt]  file {Potts/unsorted_noise.txt }; 

\nextgroupplot[ 
 width=\columnwidth, 
 height=0.6\columnwidth, 
 xmajorgrids, 
 yminorticks=true, 
 ymajorgrids, 
 yminorgrids,
  xticklabels=\empty,
  yticklabels=\empty,
  xmin=0,
  xmax = 100,
   title={(c) $\ell_1$-Potts solution on the sorted vector $\veca$},
tick label style={font=\large}
 ]

 \addplot  [color=black,
                only marks,
                mark=square*,
                mark size = 0.5pt,
                mark options={solid},
                line width=1.2pt]  file {Potts/sorted_orig.txt };

 \addplot  [color=blue,
                only marks,
                mark=star,
                mark size = 1.5pt,
                mark options={solid},
                line width=1.2pt]  file {Potts/sorted_noise.txt }; 

  \addplot  [color=red,
                only marks,
                mark=-,
                mark size = 1.8pt,
                opacity=0.7,
                line width=2.5pt]  file {Potts/sorted_potts.txt };

\nextgroupplot[ 
 width=\columnwidth, 
 height=0.6\columnwidth, 
 xmajorgrids, 
 yminorticks=true, 
 ymajorgrids, 
 yminorgrids,
 xticklabels=\empty,
  yticklabels=\empty,
  xmin=0,
  xmax = 100,
    title={(d) $\ell_1$-sorted-Potts solution}
 ]

 \addplot  [color=black,
                only marks,
                mark=square*,
                mark size = 0.5pt,
                mark options={solid},
                line width=1.2pt]  file {Potts/unsorted_orig.txt };

 \addplot  [color=blue,
                only marks,
                mark=star,
                mark size = 1.5pt,
                mark options={solid},
                line width=1.2pt]  file {Potts/unsorted_noise.txt };

  \addplot  [color=red,
                only marks,
                mark=square*,
                mark size = 0.5pt,
                opacity = 0.7,
                mark options={solid},
                line width=1.2pt] file {Potts/unsorted_potts.txt };

\end{groupplot}


\path (plotP c1r1.north west|-current bounding box.north)--
      coordinate(legendpos)
      (plotP c2r1.north east|-current bounding box.north);
\matrix[
    matrix of nodes,
    anchor=south,
    draw,
    inner sep=0.2em,
    draw
  ]at([yshift=1ex]legendpos)
  { 
\ref{plotP:GT}& \footnotesize Ground Truth &[5pt]
\ref{plotP:vec}& \footnotesize  $\veca$  &[5pt]
\ref{plotP:P}& \footnotesize  Clustering with Potts problem \\
};
\end{tikzpicture}

\caption{Clustering from cluster-generating vectors.}
\label{fig:potts}
\end{figure}
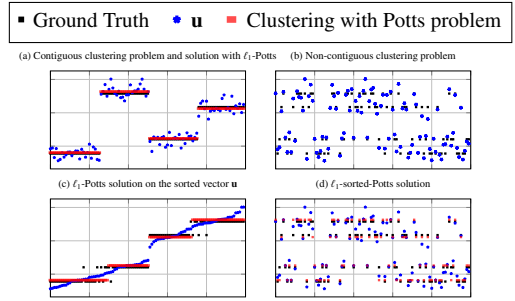

Closely related, the total variation (TV) method is based on the same formulation but with an $\ell_1$ norm instead of the $\ell_0$ semi-norm and can be seen as a convex lifting of the Potts model. 
While both these methods can be implemented efficiently in practice  \citep{friedrich2008complexity,condat2013direct}, the non-convex Potts method is more suitable for our particular application as it provides solutions with very sparse increments and thus piece-wise constant signals while the total variation method tends to accept small jumps leading in practice to different values in the signal. Consequently, in our experiments we use the $\ell_p$-Potts method to produce block-constant vectors $\clustlin$ and $\clustcol$ with few different values that can be transformed to the clustering partitions using a coordinate clustering approach. 

As for the data-related term, we choose $\ell_1$ and $\ell_2$ norms in the Potts model due to their widespread use in many real-world applications. In practice, as for regression problems, the $\ell_1$ norm tend to be more robust to noise and thus usually performs best on noisy datasets. $\ell_1$-Potts clustering is illustrated in Figure~\ref{fig:potts}a.

\subsection{Coordinates Clustering}

When the blocks in $\veca$ and $\vecb$ are non-contiguous, one cannot directly use piece-wise constant signal denoising techniques described above (see for instance the non-contiguous clustering problem of Figure~\ref{fig:potts}b). Indeed, although one may obtain distinct values for different clusters, the actual values outputted for two non-contiguous parts of the same cluster have no reason to be equal and thus they can be encoded as two different clusters. 

In order to properly solve this problem, we propose to use the \emph{$\ell_p$-sorted-Potts problem} which penalizes the increments of the \emph{sorted} vector. Denoting by $\mathfrak{s}$ the sorting operator, the $\ell_p$-sorted-Potts problem writes:
\begin{align*}
\vecc^\star = \arg\min_{\vecc \in \mathbb{R}^n}  \|\vecc-\veca\|_p^p  +  \lambda_{\text{reg}} \|D\mathfrak{s}(\vecc)\|_0 .
\end{align*}

Although this problem may appear very hard to solve, the following lemma shows that this procedure has the exact same complexity as that of sorting the vector $\veca$ and solving the $\ell_p$-Potts problem afterwards.
\begin{lemma}
Let $\mathbf{s}\in\mathbb{R}^n$ be sorted. Then, for any $p\geq 1$, $\mathbf{p}^\star = \arg\min_{\mathbf{p} \in \mathbb{R}^n}  \|\mathbf{p}-\mathbf{s}\|^p_p  +  \lambda_{\text{reg}} \|D\mathbf{p}\|_0 $ is sorted.
\end{lemma}
\begin{proof}
The proof can be found in the Appendix.
\end{proof}
Using this lemma, the solution of the {$\ell_p$-sorted-Potts problem} can be computed as: 
\begin{enumerate}
    \item Compute the sorting matrix $\mathbf{S}$ such that $\mathbf{S} \veca = \mathfrak{s}(\veca)$.
    \item Compute the solution:
    $$ \mathbf{p}^\star = \arg\min_{\mathbf{p} \in \mathbb{R}^n}  \|\mathbf{p}-\mathbf{S} \veca\|^p_p  +  \lambda_{\text{reg}} \|D\mathbf{p}\|_0.$$
    \item Desort the solution: $ \vecc^\star = \mathbf{S}^\top \mathbf{p}^\star.$
\end{enumerate}
This process is illustrated in Figures \ref{fig:potts}c and \ref{fig:potts}d. We found that the sorted Potts approach lead to a notable gain in robustness and performance compared to other methods.


\section{Numerical Illustrations}
\label{sec:EmpAn}
This section is devoted to an empirical comparison between the aforementioned approaches that can be used to produce the cluster-generating vectors. Each such vector is then post-processed using the Potts regularization described in Section \ref{sec:clust} to extract the labels. We want to point out that the objective of this comparison is to gain insights into the strengths and potential weaknesses of the different approaches for generating these vectors, as they rely on different theories, and can therefore be impacted in different ways by the structure of the input data. In addition, we aim to show the robustness of the proposed Potts schema. 

\begin{table*}[!t]
    \centering
     \caption{Mean ($\pm{}$standard deviation) of the NMI obtained for the row clustering over 100 runs on each configuration with and without noise.}
    \label{tab:cce_synthetic}
    \resizebox{0.8\textwidth}{!}{
    \begin{tabular}{cccccp{0.8cm}cccc}
    \hline
    \multirow{2}{*}{Algorithms}&\multicolumn{4}{c}{\textbf{Clean Data}}& &\multicolumn{4}{c}{\textbf{Noisy Data}}\\
    \cline{2-5}
    \cline{7-10}
    &D1&D2&D3&D4&&D1&D2&D3&D4\\
    \hline
    \CCOTGW  &$.999\pm{.002}$ &$1.00\pm{.000}$&$.819\pm{.068}$&$.925\pm{.063}$&&$.639\pm{.049}$&$.659\pm{.066}$&$.212\pm{.178}$&$.324\pm{.032}$\\
    \texttt{NMF}  &$.990\pm{.047}$&$.915\pm{.110}$&$1.00\pm{.000}$&$1.00\pm{.000}$&&$.955\pm{.077}$&$.708\pm{.125}$&$.962\pm{.075}$&$.937\pm{.027}$\\
    \texttt{Fiedler}  &$.999\pm{.000}$&$.950\pm{.093}$&$1.00\pm{.000}$&$1.00\pm{.000}$&&$1.00\pm{.000}$&$.991\pm{.043}$&$1.00\pm{.00}$&$.694\pm{.263}$\\
    \texttt{Fiedler DS} &$1.00\pm{.000}$&$1.00\pm{.000}$&$1.00\pm{.000}$&$1.00\pm{.000}$&&$1.00\pm{.000}$&$1.00\pm{.000}$&$1.00\pm{.000}$&$.900\pm{.000}$\\
    \texttt{Marginal}  &$.990\pm{.047}$&$.959\pm{.084}$&$1.00\pm{.000}$&$1.00\pm{.000}$&&$.699\pm{.045}$&$.536\pm{.08}$&$.592\pm{.011}$&$.794\pm{.018}$\\
    \hline
    \end{tabular}
    }
\end{table*}

\textit{Data} We simulate several data sets having continuous input that arise from the Gaussian latent block model, which is an extension the Gaussian mixture model, for co-clustering. We consider four scenarios by varying the number of blocks, the size of the data set, the degree of overlapping between the blocks and the proportion of each block. 
We also study the impact of noise on the structure of the cluster-generating vectors by disrupting the underlying block structure of all data sets with Gaussian white noise. 
For each scenario, we generate 100 data sets and compute the mean (and standard deviation) of the Normalized Mutual Information (NMI). One can observe that as $\lambda_{\text{reg}}$ decreases the number of constant steps, \ie, the number of clusters, increases. In this context, setting an appropriate value for $\lambda_{\text{reg}}$ can be seen as a problem of finding an appropriate number of clusters, and therefore, can be solved using the silhouette analysis \citep{Rousseeuw87}. Details regarding the generative process, the parameters and the tuning of $\lambda_{\text{reg}}$ are available in \ref{app: expe}.

\textit{Baselines} We tested two versions of the \texttt{NMF} based on Kullback-Leibler and on the Euclidean loss. As the obtained results were quite similar, we decided to omit the latter. Furthermore, as \texttt{NMF} is known to be sensitive to initialisation, we run it 100 times per generated dataset, and take the average rank-one vectors for the label detection (see Figure \ref{fig:nmf_init}). In what follows, \texttt{Fiedler} refers to the Fiedler vector obtained on the Laplacian of the similarity matrix computed from the original data. In \texttt{Fiedler DS} the Laplacian is computed from the doubly stochastic similarity matrix. This latter method can be shown to be equivalent to the first step of the approach proposed by \cite{Gorrec2019}.

\begin{figure}[!htpb]
\begin{tikzpicture}[scale=0.5]
  \centering
  \begin{axis}[
        ybar, axis on top,
        title={},
        height=8cm, width=15.5cm,
        bar width=0.45cm,
        ymajorgrids, tick align=inside,
        major grid style={draw=white},
        enlarge y limits={value=.1,upper},
        ymin=0, ymax=29.5,
        axis x line*=bottom,
        axis y line*=right,
        y axis line style={opacity=0},
        tickwidth=0pt,
        enlarge x limits=true,
        legend style={
            at={(0.5,-0.2)},
            anchor=north,
            font=\large,
            legend columns=-1,
            /tikz/every even column/.append style={column sep=0.5cm}
        },
        ylabel={Co-clustering Error (\%)},
        symbolic x coords={
           D1,D1-n,D2,D2-n,D3,D3-n,D4,D4-n},
       xtick=data,
       tick label style={font=\large}
    ]
    \addplot [draw=none, fill=gray!20] coordinates {
      (D1,0.4)
      (D1-n, 23.2) 
      (D2, 1.1) 
	  (D2-n, 29.1)      
      (D3,0.8)
      (D3-n, 21.6) 
      (D4,7.9) 
      (D4-n, 24.3) 
       };
   \addplot [draw=none,fill=gray!90] coordinates {
      (D1,0.1)
      (D1-n, 2.3) 
      (D2, 1.16) 
      (D2-n, 7.9)      
      (D3,0.2)
      (D3-n, 5.4) 
      (D4,4.5) 
     (D4-n, 8.5) 
     };
    \legend{CCOT-GW,NMF}
  \end{axis}
  \end{tikzpicture}
 \caption{Co-clustering errors obtained with {\CCOTGW} and \texttt{NMF} on all settings. ``-n'' denotes the noisy versions.}
  \label{fig:cce}
\end{figure}
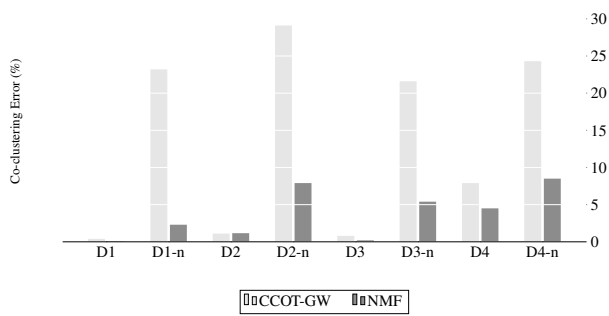

\textit{Results} We summarize the obtained results in Table \ref{tab:cce_synthetic}. From them, we observe that all methods perform equally well on the clean data with no differences that can be considered significant. This shows that all learning algorithms manage to identify the underlying block structure whether they proceed by analyzing the spectral information or the statistical properties. The situation changes when we apply these algorithms on the noisy data where important differences between the considered approaches can be observed. These differences highlight the high noise robustness of both \texttt{Fiedler} methods followed by our proposed NMF formulation. This, however, is much less the case of the {\CCOTGW} method whose performance drops significantly. One possible explanation to this is that \texttt{Fiedler} methods can be impacted by local noise inside the generated clusters while maintaining high robustness to global noise. Finally, as our proposed \texttt{NMF} formulation and {\CCOTGW} naturally provide a solution for the co-clustering problem, \ie, a partition for both the rows and the columns of the input data, we report the co-clustering errors for both methods in Figure \ref{fig:cce} for all settings. From it, we can note that \texttt{NMF} compares favorably to {\CCOTGW} in terms of the obtained performance and has a much lower computational complexity. 

\section{Conclusion and Perspectives}
In this paper, we formalized a rank-one partitioning framework and showed that two different learning frameworks and the algorithms associated with them fall into the introduced definition. Based in this formalization, we proposed a regularization strategy that can be used in conjunction with any such method and validated its usefulness through experimental evaluations. An important future perspective of our work would be to understand what supplementary information is provided by rank-$k$ decomposition as well as to consider it as a hierarchical model of rank-one methods that provide different views for the same mode of a data matrix. 

\bibliography{refs}

\begin{thebibliography}{32}
\expandafter\ifx\csname natexlab\endcsname\relax\def\natexlab#1{#1}\fi
\providecommand{\url}[1]{\texttt{#1}}
\providecommand{\href}[2]{#2}
\providecommand{\path}[1]{#1}
\providecommand{\DOIprefix}{doi:}
\providecommand{\ArXivprefix}{arXiv:}
\providecommand{\URLprefix}{URL: }
\providecommand{\Pubmedprefix}{pmid:}
\providecommand{\doi}[1]{\href{http://dx.doi.org/#1}{\path{#1}}}
\providecommand{\Pubmed}[1]{\href{pmid:#1}{\path{#1}}}
\providecommand{\bibinfo}[2]{#2}
\ifx\xfnm\relax \def\xfnm[#1]{\unskip,\space#1}\fi
\bibitem[{Basseville and Nikiforov(1993)}]{basseville1993detection}
\bibinfo{author}{Basseville, M.}, \bibinfo{author}{Nikiforov, I.V.},
  \bibinfo{year}{1993}.
\newblock \bibinfo{title}{Detection of abrupt changes: theory and application}.
  volume \bibinfo{volume}{104}.
\newblock \bibinfo{publisher}{Prentice Hall Englewood Cliffs}.
\bibitem[{Benamou et~al.(2015)Benamou, Carlier, Cuturi, Nenna and
  Peyr{\'e}}]{2015-Benamou-Bregman}
\bibinfo{author}{Benamou, J.D.}, \bibinfo{author}{Carlier, G.},
  \bibinfo{author}{Cuturi, M.}, \bibinfo{author}{Nenna, L.},
  \bibinfo{author}{Peyr{\'e}, G.}, \bibinfo{year}{2015}.
\newblock \bibinfo{title}{{Iterative Bregman Projections for Regularized
  Transportation Problems}}.
\newblock \bibinfo{journal}{{SIAM Journal on Scientific Computing}}
  \bibinfo{volume}{2}, \bibinfo{pages}{A1111--A1138}.
\bibitem[{Brault and Channarond(2017)}]{brault_17}
\bibinfo{author}{Brault, V.}, \bibinfo{author}{Channarond, A.},
  \bibinfo{year}{2017}.
\newblock \bibinfo{title}{Fast and consistent algorithm for the latent block
  model}.
\newblock \bibinfo{journal}{CoRR} \bibinfo{volume}{abs/1610.09005}.
\bibitem[{Channarond et~al.(2012)Channarond, Daudin and Robin}]{channarond2012}
\bibinfo{author}{Channarond, A.}, \bibinfo{author}{Daudin, J.J.},
  \bibinfo{author}{Robin, S.}, \bibinfo{year}{2012}.
\newblock \bibinfo{title}{Classification and estimation in the stochastic
  blockmodel based on the empirical degrees}.
\newblock \bibinfo{journal}{Electron. J. Statist.} \bibinfo{volume}{6},
  \bibinfo{pages}{2574--2601}.
\bibitem[{Collins et~al.(2001)Collins, Dasgupta and Schapire}]{Collins:2001}
\bibinfo{author}{Collins, M.}, \bibinfo{author}{Dasgupta, S.},
  \bibinfo{author}{Schapire, R.E.}, \bibinfo{year}{2001}.
\newblock \bibinfo{title}{A generalization of principal component analysis to
  the exponential family}, in: \bibinfo{booktitle}{NIPS}, pp.
  \bibinfo{pages}{617--624}.
\bibitem[{Condat(2013)}]{condat2013direct}
\bibinfo{author}{Condat, L.}, \bibinfo{year}{2013}.
\newblock \bibinfo{title}{A direct algorithm for 1-d total variation
  denoising}.
\newblock \bibinfo{journal}{IEEE Signal Processing Letters}
  \bibinfo{volume}{20}, \bibinfo{pages}{1054--1057}.
\bibitem[{Cuturi(2013)}]{conf/nips/Cuturi13}
\bibinfo{author}{Cuturi, M.}, \bibinfo{year}{2013}.
\newblock \bibinfo{title}{Sinkhorn distances: Lightspeed computation of optimal
  transport.}, in: \bibinfo{booktitle}{NIPS}, pp. \bibinfo{pages}{2292--2300}.
\bibitem[{Ding et~al.(2001)Ding, He and Zha}]{DingHZ01}
\bibinfo{author}{Ding, C.H.Q.}, \bibinfo{author}{He, X.}, \bibinfo{author}{Zha,
  H.}, \bibinfo{year}{2001}.
\newblock \bibinfo{title}{A spectral method to separate disconnected and
  nearly-disconnected web graph components}, in: \bibinfo{booktitle}{KDD}, pp.
  \bibinfo{pages}{275--280}.
\bibitem[{Ding et~al.(2010)Ding, Li and Jordan}]{Ding:2010:CSM:1687044.1687110}
\bibinfo{author}{Ding, C.H.Q.}, \bibinfo{author}{Li, T.},
  \bibinfo{author}{Jordan, M.I.}, \bibinfo{year}{2010}.
\newblock \bibinfo{title}{Convex and semi-nonnegative matrix factorizations}.
\newblock \bibinfo{journal}{IEEE Trans. Pattern Anal. Mach. Intell.}
  \bibinfo{volume}{32}, \bibinfo{pages}{45--55}.
\bibitem[{F{\'e}votte and Idier(2011)}]{fevotte2011algorithms}
\bibinfo{author}{F{\'e}votte, C.}, \bibinfo{author}{Idier, J.},
  \bibinfo{year}{2011}.
\newblock \bibinfo{title}{Algorithms for nonnegative matrix factorization with
  the $\beta$-divergence}.
\newblock \bibinfo{journal}{Neural computation} \bibinfo{volume}{23},
  \bibinfo{pages}{2421--2456}.
\bibitem[{Fiedler(1973)}]{fiedler73}
\bibinfo{author}{Fiedler, M.}, \bibinfo{year}{1973}.
\newblock \bibinfo{title}{Algebraic connectivity of graphs}.
\newblock \bibinfo{journal}{Czechoslovak Mathematical Journal}
  \bibinfo{volume}{23}, \bibinfo{pages}{298--305}.
\bibitem[{Friedrich et~al.(2008)Friedrich, Kempe, Liebscher and
  Winkler}]{friedrich2008complexity}
\bibinfo{author}{Friedrich, F.}, \bibinfo{author}{Kempe, A.},
  \bibinfo{author}{Liebscher, V.}, \bibinfo{author}{Winkler, G.},
  \bibinfo{year}{2008}.
\newblock \bibinfo{title}{Complexity penalized m-estimation: fast computation}.
\newblock \bibinfo{journal}{Journal of Computational and Graphical Statistics}
  \bibinfo{volume}{17}, \bibinfo{pages}{201--224}.
\bibitem[{Kantorovich(1942)}]{kantorovich}
\bibinfo{author}{Kantorovich, L.}, \bibinfo{year}{1942}.
\newblock \bibinfo{title}{On the translocation of masses}, in:
  \bibinfo{booktitle}{C.R. (Doklady) Acad. Sci. URSS(N.S.)}, pp.
  \bibinfo{pages}{199--201}.
\bibitem[{Knight(2008)}]{knight_08}
\bibinfo{author}{Knight, P.A.}, \bibinfo{year}{2008}.
\newblock \bibinfo{title}{The sinkhorn–knopp algorithm: Convergence and
  applications}.
\newblock \bibinfo{journal}{SIAM Journal on Matrix Analysis and Applications}
  \bibinfo{volume}{30}, \bibinfo{pages}{261--275}.
\bibitem[{Koren et~al.(2009)Koren, Bell and
  Volinsky}]{Koren:2009:MFT:1608565.1608614}
\bibinfo{author}{Koren, Y.}, \bibinfo{author}{Bell, R.},
  \bibinfo{author}{Volinsky, C.}, \bibinfo{year}{2009}.
\newblock \bibinfo{title}{Matrix factorization techniques for recommender
  systems}.
\newblock \bibinfo{journal}{Computer} \bibinfo{volume}{42},
  \bibinfo{pages}{30--37}.
\bibitem[{Laclau et~al.(2017)Laclau, Redko, Matei, Bennani and
  Brault}]{laclau_17}
\bibinfo{author}{Laclau, C.}, \bibinfo{author}{Redko, I.},
  \bibinfo{author}{Matei, B.}, \bibinfo{author}{Bennani, Y.},
  \bibinfo{author}{Brault, V.}, \bibinfo{year}{2017}.
\newblock \bibinfo{title}{Co-clustering through optimal transport}, in:
  \bibinfo{booktitle}{{ICML}}, pp. \bibinfo{pages}{1955--1964}.
\bibitem[{Le~Gorrec et~al.(2019)Le~Gorrec, Mouysset, Duff, Knight and
  Ruiz}]{Gorrec2019}
\bibinfo{author}{Le~Gorrec, L.}, \bibinfo{author}{Mouysset, S.},
  \bibinfo{author}{Duff, I.}, \bibinfo{author}{Knight, P.},
  \bibinfo{author}{Ruiz, D.}, \bibinfo{year}{2019}.
\newblock \bibinfo{title}{{Uncovering Hidden Block Structure for Clustering}},
  in: \bibinfo{booktitle}{{European Conference on Machine Learning and
  Principles and Practice of Knowledge Discovery in Databases (ECML-PKDD)}},
  \bibinfo{publisher}{Springer}.
\bibitem[{Lee and Seung()}]{lee99}
\bibinfo{author}{Lee, D.D.}, \bibinfo{author}{Seung, H.S.},
  \bibinfo{year}{1999}.
\newblock \bibinfo{title}{Learning the parts of objects by nonnegative matrix
  factorization}.
\newblock \bibinfo{journal}{Nature} \bibinfo{volume}{401},
  \bibinfo{pages}{788--791}.
\bibitem[{Little and Jones(2011)}]{little2011generalized}
\bibinfo{author}{Little, M.A.}, \bibinfo{author}{Jones, N.S.},
  \bibinfo{year}{2011}.
\newblock \bibinfo{title}{Generalized methods and solvers for noise removal
  from piecewise constant signals. i. background theory}, in:
  \bibinfo{booktitle}{Proc. R. Soc. A}, pp. \bibinfo{pages}{3088--3114}.
\bibitem[{Luxburg(2007)}]{Luxburg:2007:TSC:1288822.1288832}
\bibinfo{author}{Luxburg, U.}, \bibinfo{year}{2007}.
\newblock \bibinfo{title}{A tutorial on spectral clustering}.
\newblock \bibinfo{journal}{Statistics and Computing} \bibinfo{volume}{17},
  \bibinfo{pages}{395--416}.
\bibitem[{Matei and Meignen(2012)}]{MatMei}
\bibinfo{author}{Matei, B.}, \bibinfo{author}{Meignen, S.},
  \bibinfo{year}{2012}.
\newblock \bibinfo{title}{Nonlinear cell-average multiscale signal
  representations: Application to signal denoising}.
\newblock \bibinfo{journal}{Signal Processing} \bibinfo{volume}{92},
  \bibinfo{pages}{2738--2746}.
\bibitem[{Meila and Shi(2001)}]{Meila01learningsegmentation}
\bibinfo{author}{Meila, M.}, \bibinfo{author}{Shi, J.}, \bibinfo{year}{2001}.
\newblock \bibinfo{title}{Learning segmentation by random walks}, in:
  \bibinfo{booktitle}{In Advances in Neural Information Processing Systems},
  pp. \bibinfo{pages}{873--879}.
\bibitem[{Mirzal(2011)}]{mirzal11}
\bibinfo{author}{Mirzal, A.}, \bibinfo{year}{2011}.
\newblock \bibinfo{title}{Clustering and latent semantic indexing aspects of
  the nonnegative matrix factorization}.
\newblock \bibinfo{journal}{CoRR} \bibinfo{volume}{abs/1112.4020}.
\bibitem[{Monge(1781)}]{monge_81}
\bibinfo{author}{Monge, G.}, \bibinfo{year}{1781}.
\newblock \bibinfo{title}{M\'emoire sur la th\'eorie des d\'eblais et des
  remblais}.
\newblock \bibinfo{journal}{Histoire de l'Acad\'emie Royale des Sciences} ,
  \bibinfo{pages}{666--704}.
\bibitem[{Page et~al.(1998)Page, Brin, Motwani and Winograd}]{Pageetal98}
\bibinfo{author}{Page, L.}, \bibinfo{author}{Brin, S.},
  \bibinfo{author}{Motwani, R.}, \bibinfo{author}{Winograd, T.},
  \bibinfo{year}{1998}.
\newblock \bibinfo{title}{The pagerank citation ranking: Bringing order to the
  web}, in: \bibinfo{booktitle}{WWW}, pp. \bibinfo{pages}{161--172}.
\bibitem[{Patrikainen and Meila(2006)}]{Patrikainen06}
\bibinfo{author}{Patrikainen, A.}, \bibinfo{author}{Meila, M.},
  \bibinfo{year}{2006}.
\newblock \bibinfo{title}{Comparing subspace clusterings}.
\newblock \bibinfo{journal}{IEEE Transactions on Knowledge and Data
  Engineering} \bibinfo{volume}{18}, \bibinfo{pages}{902--916}.
\bibitem[{Peyr{\'{e}} et~al.(2016)Peyr{\'{e}}, Cuturi and
  Solomon}]{DBLP:conf/icml/PeyreCS16}
\bibinfo{author}{Peyr{\'{e}}, G.}, \bibinfo{author}{Cuturi, M.},
  \bibinfo{author}{Solomon, J.}, \bibinfo{year}{2016}.
\newblock \bibinfo{title}{Gromov-wasserstein averaging of kernel and distance
  matrices}, in: \bibinfo{booktitle}{ICML}, pp. \bibinfo{pages}{2664--2672}.
\bibitem[{Rousseeuw(1987)}]{Rousseeuw87}
\bibinfo{author}{Rousseeuw, P.}, \bibinfo{year}{1987}.
\newblock \bibinfo{title}{Silhouettes: A graphical aid to the interpretation
  and validation of cluster analysis}.
\newblock \bibinfo{journal}{J. Comput. Appl. Math.} \bibinfo{volume}{20},
  \bibinfo{pages}{53--65}.
\bibitem[{Shi and Malik(2000)}]{Shi:2000:NCI:351581.351611}
\bibinfo{author}{Shi, J.}, \bibinfo{author}{Malik, J.}, \bibinfo{year}{2000}.
\newblock \bibinfo{title}{Normalized cuts and image segmentation}.
\newblock \bibinfo{journal}{IEEE Trans. Pattern Anal. Mach. Intell.}
  \bibinfo{volume}{22}, \bibinfo{pages}{888--905}.
\bibitem[{Sinkhorn and Knopp(1967)}]{sinknopp_67}
\bibinfo{author}{Sinkhorn, R.}, \bibinfo{author}{Knopp, P.},
  \bibinfo{year}{1967}.
\newblock \bibinfo{title}{Concerning nonnegative matrices and doubly stochastic
  matrices}.
\newblock \bibinfo{journal}{Pacific Journal of Mathematics}
  \bibinfo{volume}{21}, \bibinfo{pages}{343--348}.
\bibitem[{Snijders and Nowicki(1997)}]{Snijders1997}
\bibinfo{author}{Snijders, T.A.}, \bibinfo{author}{Nowicki, K.},
  \bibinfo{year}{1997}.
\newblock \bibinfo{title}{Estimation and prediction for stochastic blockmodels
  for graphs with latent block structure}.
\newblock \bibinfo{journal}{Journal of Classification} \bibinfo{volume}{14},
  \bibinfo{pages}{75--100}.
\bibitem[{Weinmann et~al.(2015)Weinmann, Storath and Demaret}]{weinmann20151}
\bibinfo{author}{Weinmann, A.}, \bibinfo{author}{Storath, M.},
  \bibinfo{author}{Demaret, L.}, \bibinfo{year}{2015}.
\newblock \bibinfo{title}{The $l^1$-potts functional for robust jump-sparse
  reconstruction}.
\newblock \bibinfo{journal}{SIAM Journal on Numerical Analysis}
  \bibinfo{volume}{53}, \bibinfo{pages}{644--673}.

\end{thebibliography}
\newpage
\appendix

\section{Recalls about Optimal Transport}
\label{sec:ot}
Optimal transport \citep{monge_81} is the branch of mathematics that considers the problem of finding a mapping transporting one probability measure to another one in a way that minimizes the transportation cost.

In order to introduce it, let us consider two discrete probability measures:
$$\measa := \sum_{i=1}^m a_i \delta_{\lin_i} ~~ \text{ and } ~~ \measb := \sum_{j=1}^n b_j \delta_{\col_j}$$
defined as weighted sums of Diracs over two point sets $\{\lin_i \in \mathbb{R}^{d_{\lin}} \}_{i=1}^m$ and $\{\col_j \in \mathbb{R}^{d_{\col}} \}_{j=1}^n$. We denote by $\mathbf{a}$ and $\mathbf{b}$ the corresponding vectors of probabilities of the two distributions, belonging to the simplex of size-$m$ and size-$n$, respectively.

To transport a discrete measure $\measa$ to another discrete measure $\measb$, the relaxation of the original transportation problem proposed by Kantorovich  \citep{kantorovich} consists in finding a coupling matrix $\coupl \in \mathbb{R}^{m\times n}$ (instead of a mapping evoked above) such that $\coupl_{i,j}$ is defined as the fraction of mass transported from the $i$-th bin of the source distribution $\measa$ (i.e. $\lin_i$) to the $j$-th bin of the target distribution $\measb$ (i.e. $\col_j$). More precisely, the $i$-th row of $\coupl$ corresponds to the proportions of source bin $i$ splitting its mass $a_i$ across the target bins. This latter condition means that the sum of all entries of $\coupl$ is equal to one. The set of admissible couplings thus writes:
\begin{align*}
    \Pi(\measa,\measb):=\left\{ \coupl \in \mathbb{R}_{+}^{m\times n} : \coupl \one_n = \mathbf{a} , \coupl^\top \one_m = \mathbf{b}  \right\},
\end{align*}
where $\one_n$ is the size-$n$ vector of ones. 

To find an optimal coupling among the admissible ones, one has to define a cost matrix $\matM \in \mathbb{R}_{+}^{m\times n} $ that models the cost of moving source bin $i$ to target bin $j$ represented in our case by $\lin_i$ and $\col_j$, respectively. For instance, if $d_{\lin} =d_{\col} $\footnote{Note that when this is not the case, one may use the Gromov-Wasserstein distance defined for metric-measure spaces of different size. We refer the reader to \citep{DBLP:conf/icml/PeyreCS16} for further details.}
, a natural choice for $\matM$ is the squared Euclidean distance, \ie~ $M_{i,j} = \|\lin_i - \col_j \|_2^2$. Once this cost matrix is defined, the Kantorovich optimal transportation problem from $\measa$ to $\measb$ writes:
\begin{align*}
    \min_{\coupl\in\Pi(\measa,\measb)} \langle \matM , \coupl \rangle_F,
\end{align*}
where $\langle \cdot , \cdot \rangle_F$ denotes the Frobenius scalar product. 

Despite the appealing nature of the problem, its applicability remained limited in practice due to its high complexity. Indeed, the underlying optimization problem can be formulated as a Linear Program with worst-case complexity $\mathcal{O}(n^3\log(n))$. To tackle this drawback, \citep{conf/nips/Cuturi13} proposed to add a regularization to promote couplings that are close to the trivial coupling $\mathbf{a}\mathbf{b}^\top\in \Pi(\measa,\measb)$ in the sense of the Kullback-Liebler divergence:
\begin{align*}
    \KL(\coupl || \mathbf{a}\mathbf{b}^\top) &:= \sum_{i,j=1}^{m,n} \coupl_{i,j} \log\left( \frac{\coupl_{i,j}}{a_i b_j} \right) - \gamma_{i,j} + a_i b_j\\
    &= \sum_{i,j=1}^{m,n} \coupl_{i,j} \log ( \coupl_{i,j} ) - \sum_{i=1}^{m} a_{i} \log (a_{i} )  - \sum_{i=1}^{n} b_{i} \log (b_{i} )\\
    &= -E(\coupl) + E(\mathbf{a}) + E(\mathbf{b}), 
\end{align*}
where we use the fact that the full, row, and column sums of an admissible coupling $\coupl\! \in \!\Pi(\measa,\measb)$ are equal to $1$,  $\mathbf{a}$, and $\mathbf{b}$, respectively and where $E(\mathbf{a})\! :=\! - \sum_{i=1}^{m} a_{i} \log (a_{i} ) $ stands for the entropy of a probability vector. Finally, dropping the terms independent of $\coupl$, the regularized optimal transport from $\measa$ to $\measb$ writes:
\begin{align}
\label{eq:regot}
    \min_{\coupl\in\Pi(\measa,\measb)} \langle \matM , \coupl \rangle_F - \varepsilon E(\coupl),
\end{align}
where $\varepsilon\geq0$ is a regularization parameter. This regularization allows to obtain smoother and more numerically stable solutions compared to the original case  \citep{2015-Benamou-Bregman}. Indeed, Sinkhorn's theorem \citep{sinknopp_67} tells us that for $\varepsilon>0$, \eqref{eq:regot} has a unique solution that can be obtained by left and right scaling of the Gibbs kernel $\e^{-\matM/\varepsilon}$ (where the exponential is taken element-wise) to the prescribed sums of the admissible couplings:
\begin{align*}
   \coupl_{\varepsilon}^\star (\measa,\measb) &:=  \arg\min_{\coupl\in\Pi(\measa,\measb)} \langle \matM , \coupl \rangle_F - \varepsilon E(\coupl)=  \text{diag}(\veca ) \e^{-\matM/\varepsilon}\text{diag}(\vecb),
\end{align*}
where $\veca$ and $\vecb$ are two non-negative scaling vectors uniquely defined up to a multiplicative factor, that can be efficiently computed using Sinkhorn's algorithm. 

\section{Proof of Lemma}
\label{app:proof}
\addtocounter{lemma}{-1}
\begin{lemma}
Let $\mathbf{s}\in\mathbb{R}^n$ be sorted. Then, for any $p\geq 1$, $\mathbf{p}^\star = \arg\min_{\mathbf{p} \in \mathbb{R}^n}  \|\mathbf{p}-\mathbf{s}\|^p_p  +  \lambda_{\text{reg}} \|D\mathbf{p}\|_0 $ is sorted.
\end{lemma}
\begin{proof}
The proof relies on an induction over the number of jumps (or similarly the number of different coordinate values) in the solution. If $\mathbf{p}^\star$ has $0$ jumps, it has only one value, and is thus sorted. 

Then, if $\mathbf{p}^\star$ has $1$ jump at position $i\in[1,n-1]$ (identified as the last index before the jump), then it takes two values $p_1^\star, p_2^\star \in \mathbb{R}$. The {problem} can then be decomposed into two real-valued problems $p_1^\star := \argmin_{p_1\in\mathbb{R}} \sum_{j\leq i} | \mathbf{s}_i - p_1 |^p $ and  $p_2^\star := \argmin_{p_2\in\mathbb{R}} \sum_{j> i} | \mathbf{s}_i - p_2 |^p $. If $p=1$, $p_1^\star$ is simply the median of the values before the jump; if $p=2$,  $p_1^\star$ is the average of the same values. For both cases, if $\mathbf{s}$ is sorted, $p_1^\star < p_2^\star$ and $\mathbf{p}$ is also sorted. 

Now, let us assume a sorted vector $\mathbf{p}^\star$ with $N$ jumps. Then, one can prove that a vector $\mathbf{p}^\star$ with $N+1$ is sorted by noting that the values before the $N$-th jump are sorted from the assumption and use the argument for $1$ jump to conclude that the values after the $N$-th jump are sorted  and greater than the ones before it, concluding the induction. 
\end{proof}
\section{Experimental Analysis}
\label{app: expe}

\paragraph{Generative process} The Gaussian Latent Block Model postulates that a data matrix is drawn from the following generative procedure.
\begin{itemize}
\item Generating $\clustlin$ according to a Multinomial distribution $\mathcal{M}(1;\pi_1, \cdots, \pi_{k_r})$, where $\pi$'s represents the a priori proportion of rows in each cluster;
\item Generating $\clustcol$ according to a Multinomial distribution $\mathcal{M}(1;\tau_1, \cdots, \tau_{k_c})$ where $\tau$'s represents the a priori proportion of columns in each cluster;.
\item Generating $A$ with for each $j\in\{1,\ldots,n\}$: $\forall A, A_{ij}\sim \mathcal{N}(\alpha_{\clustlin_i\clustcol_j}) $
\end{itemize}

\paragraph{Simulated Data }
Table \ref{aaai17:tab:data_description} presents the details of all data configurations. 
\begin{table}[!h]
\renewcommand\thetable{1}
\caption{Size ($m\times n$), number of co-clusters ($k_r$, $k_c$), degree of overlapping ([+] for well-separated and [++] for ill-separated co-clusters) and the proportions of co-clusters for simulated data sets.}
\label{aaai17:tab:data_description}
\centering
\resizebox{0.4\textwidth}{!}{
\begin{tabular}{lccccl}
\hline
Data set&$n\times m$&$g\times k$&Overlapping&Proportions\\
\hline
D1&$600\times300$&$3\times 3$&[+]&Equal\\
D2&$600\times300$&$3\times 3$&[+]&Unequal\\
D3&$300\times200$&$2\times 4$&[++]&Equal\\
D4&$300\times300$&$5\times4$&[++]&Unequal\\
\hline
\end{tabular}
}
\end{table}
We used two metrics for measuring the performance of the difference approaches for the clustering and the co-clustering tasks. The normalized mutual information and the co-clustering error. 

The co-clustering error (CCE) \cite{Patrikainen06} is defined as follows
\begin{equation*}
\resizebox{0.8\hsize}{!}{
   $
\text{CCE}((\clustlin,\clustcol),(\hatclustlin,\hatclustcol))\!=\!e(\clustlin,\hatclustlin)\!+\! e(\clustcol,\hatclustcol)\!-\!e(\clustlin,\hatclustlin) e(\clustcol,\hatclustcol),$
}
\end{equation*}
where $\hatclustlin$ and $\hatclustcol$ are the partitions of instances and variables estimated by the algorithm; $\clustlin$ and $\clustcol$ are the true partitions and $e(\clustlin,\hatclustlin)$ (resp. $e(\clustcol,\hatclustcol)$) denotes the error rate, i.e., the proportion of misclassified instances (resp. features).

Finally, as stated in the Experimental Section we propose to add a perturbation to the generated dataset to understand the impact of noise on the generated vectors. An illustration of such effect is shown in Figure \ref{fig:noise}.

\begin{figure}[!htpb]
\centering
 \begin{tabular}[t]{cc}
\begin{tikzpicture}[scale=0.4]
\begin{axis}[ 
 width=\columnwidth, 
 height=0.75\columnwidth, 
 xmajorgrids, 
 yminorticks=true, 
 ymajorgrids, 
 yminorgrids,
 ymin = 2.4,
 ymax=2.9,
 xmin = 1,
 xmax = 300,
   ylabel={~},
 xlabel = {~},
  xticklabels=\empty,
  yticklabels=\empty,
legend style={at={(0,1)},anchor=north west, font=\large},
tick label style={font=\large},
legend image post style={scale=3.5}
 ]

 \addplot  [color=green!50!black,
                only marks,
                mark=*,
                mark size = 0.8pt,
                mark options={solid},
                line width=0.5pt]  file {u_nmf.txt }; 
 \addlegendentry{$\veca$};

  \addplot  [color=blue,
                only marks,
                mark=star,
                mark size = 0.8pt,
                mark options={solid},
                line width=0.5pt]    file {u_nmf_noise.txt}; 
 \addlegendentry{$\veca$ noise };

\end{axis}
\end{tikzpicture}
&
\begin{tikzpicture}[scale=0.4]

\begin{axis}[ 
 width=\columnwidth, 
 height=0.75\columnwidth, 
 xmajorgrids, 
 yminorticks=true, 
 ymajorgrids, 
 yminorgrids,
 ymin = 2.4,
 ymax=2.9,
 xmin = 1,
 xmax = 300,
  ylabel={~},
 xlabel = {~},
xticklabels={},
yticklabels={},
extra x ticks={0},
extra x tick labels={~},
extra y ticks={0},
extra y tick labels={~},
legend style={at={(0,1)},anchor=north west, font=\large},
tick label style={font=\large},
legend image post style={scale=3.5}
 ]

 \addplot   [color=green!50!black,
                only marks,
                mark=*,
                mark size = 0.8pt,
                mark options={solid},
                line width=0.5pt]  file {v_nmf.txt}; 
 \addlegendentry{$\vecb$};

  \addplot [color=blue,
                only marks,
                mark=star,
                mark size = 0.8pt,
                mark options={solid},
                line width=0.5pt]  file {v_nmf_noise.txt}; 
 \addlegendentry{$\vecb$ noise };
\end{axis}
\end{tikzpicture}
\end{tabular}
\caption{Sorted vectors $\veca$ (left) and $\vecb$ (right) obtained using the \NMF on D4 and D4 data sets with white noise.}
    \label{fig:noise}
\end{figure}

\begin{figure}[!htpb]
    \centering
    \subfloat{\includegraphics[width=0.2\textwidth]{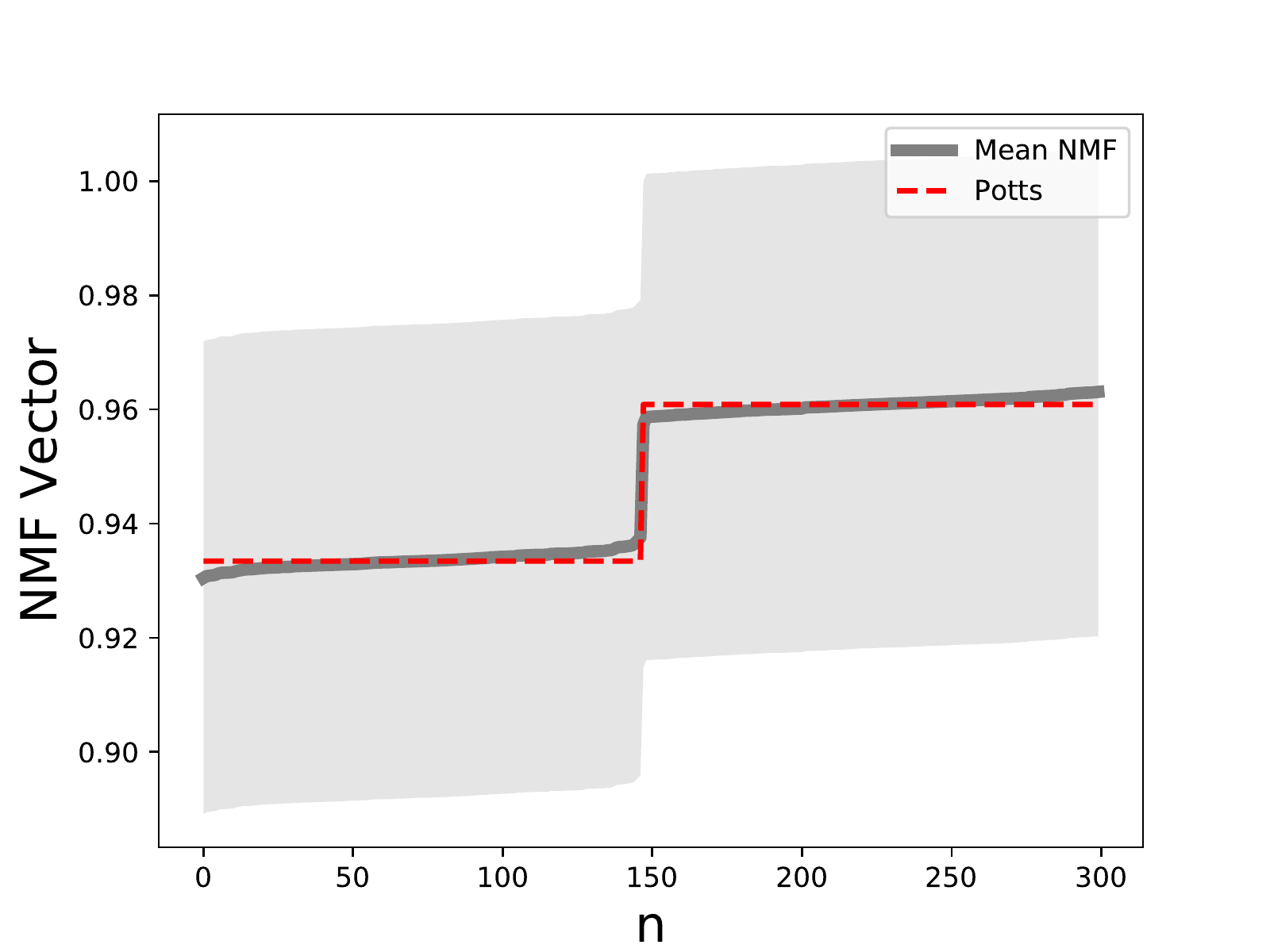}}
    \subfloat{\includegraphics[width=0.2\textwidth]{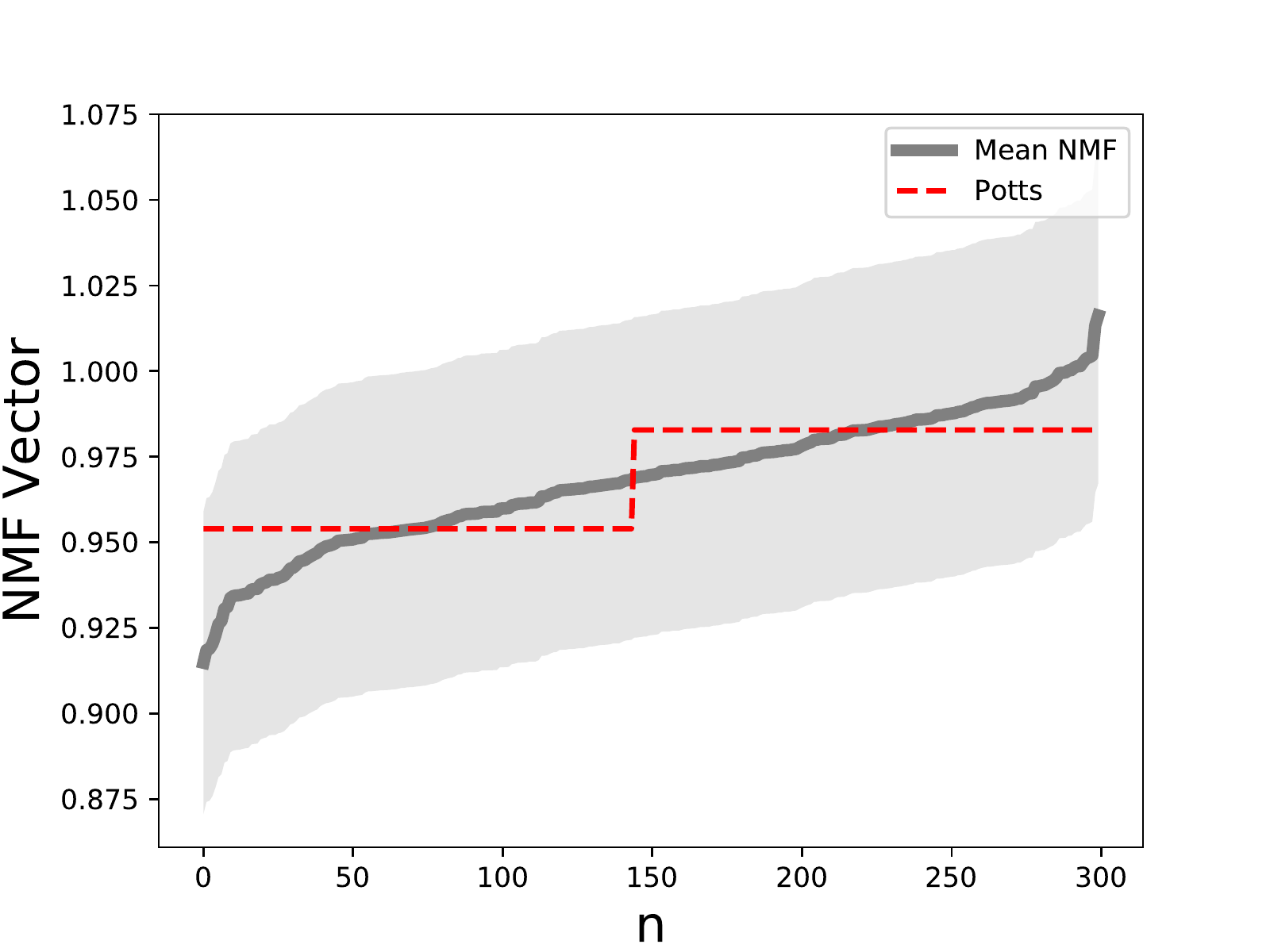}}
    \caption{Average vector (dark grey line) generated by the NMF algorithm over 100 trials. The light grey area represents the standard deviation and the red dot line is the denoised signal obtained by Potts.}
    \label{fig:nmf_init}
\end{figure}

\paragraph{Multiple initializations for NMF}
As we know that the performance of NMF are strongly impacted by the initialization of the low-dimensional matrices (in our case vectors), we propose to run this latter 100 times and to take the average vector as the one on which we apply the Potts regularization. An illustration of this point is shown in Figure \ref{fig:nmf_init}.

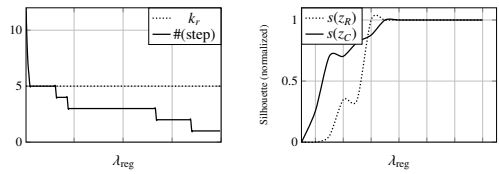
\begin{figure}[!htpb]
\centering
 \begin{tabular}[t]{cc}
\begin{tikzpicture}[scale=0.4]

\begin{axis}[ 
 width=\columnwidth, 
 height=0.75\columnwidth, 
 xmajorgrids, 
 yminorticks=true, 
 ymajorgrids, 
 yminorgrids,
 ylabel={},
 xlabel = {$\lambda_{\text{reg}}$},
 ymin = 0,
 ymax=12,
 xmin = 1,
 xmax = 200,
 xticklabels={},
legend style={at={(1,1)},anchor=north east, font=\Large},
tick label style={font=\large},
x label style={at={(axis description cs:0.5,0.05)},anchor=north, font=\Large},
 ]
 \addplot  [color=black,
                dotted,
                mark=none,
                mark options={solid},
                smooth,
                line width=1.2pt]  file {true.txt}; 
 \addlegendentry{$k_r$};

  \addplot  [color=black,
                mark=none,
                mark options={solid},
                smooth,
                line width=1.2pt]  file {card_u.txt}; 
 \addlegendentry{\#(step)};
\end{axis}
\end{tikzpicture}
&\begin{tikzpicture}[scale=0.4]

\begin{axis}[ 
 width=\columnwidth, 
 height=0.75\columnwidth, 
 xmajorgrids, 
 yminorticks=true, 
 ymajorgrids, 
 yminorgrids,
 ylabel={Silhouette (normalized)},
 xlabel = {$\lambda_{\text{reg}}$},
 ymin = 0,
 ymax=1.1,
 xmin = 1,
 xmax = 15,
 xticklabels={},
legend style={at={(0,1)},anchor=north west, font=\Large},
tick label style={font=\large},
x label style={at={(axis description cs:0.5,0.05)},anchor=north, font=\Large},
 ]
 \addplot  [color=black,
                dotted,
                mark=none,
                mark options={solid},
                smooth,
                line width=1.2pt]  file {sil_z.txt }; 
 \addlegendentry{$s(z_R)$};

  \addplot  [color=black,
                mark=none,
                mark options={solid},
                smooth,
                line width=1.2pt]  file {sil_w.txt}; 
 \addlegendentry{$s(z_C)$};
\end{axis}
\end{tikzpicture}
\\
\end{tabular}
\caption{ (left) Evolution of the number of unique values, \ie the number of steps in $u$ with varying $\lambda_{reg}$; (right) evolution of the mean silhouette obtained for instances ($s_z$) and variables ($s_w$) as a function of the regularization parameter $\varepsilon$ for D4 data set with white noise.}
    \label{fig:nbCluster}
\end{figure}

\paragraph{Tuning the number of clusters through $\lambda_{\text{reg}}$}
Finding $\lambda_{\text{reg}}$ boils down to finding the appropriate number of clusters, which is a very challenging task in unsupervised learning. To this end, we compute, the mean silhouette coefficient obtained for different increasing values of $\lambda_{\text{reg}}$ and stop the procedure when its value stabilizes or starts decreasing. Figure \ref{fig:nbCluster} (right) shows the evolution of the silhouette index on the noisy version of D4 when $\varepsilon$ is varying in the interval $[0.1,1.5]$. 

\end{document}